\documentclass[letterpaper, 10 pt, conference]{ieeeconf}
\IEEEoverridecommandlockouts
\def\BibTeX{{\rm B\kern-.05em{\sc i\kern-.025em b}\kern-.08em
    T\kern-.1667em\lower.7ex\hbox{E}\kern-.125emX}}

\usepackage[font=small,skip=0pt]{subcaption}
\usepackage[normalem]{ulem}
\usepackage[colorinlistoftodos]{todonotes}
\usepackage{amsmath,amssymb,amsfonts}
\usepackage{algorithmic}
\usepackage{graphicx}
\usepackage{textcomp}
\usepackage{xcolor}
\usepackage{tikz}
\usetikzlibrary{arrows,shapes,backgrounds,patterns,fadings,decorations.pathreplacing,decorations.pathmorphing}
\tikzset{>=stealth'}
\usetikzlibrary{arrows}
\usepackage{wrapfig}
\usepackage{placeins}
\usepackage{color, colortbl}
\usepackage[font=footnotesize]{caption}
\definecolor{Gray}{gray}{0.9}

\usepackage{commands}

\newcommand{\modelparams}{\mathbf{w}}
\newcommand{\vaeencoder}{\phi}
\newcommand{\vaedecoder}{\theta}

\newcommand{\ptrain}{p_\text{train}}
\newcommand{\ptest}{p_\text{test}}
\newcommand{\datatrain}{\data_{\text{train}}}

\newcommand{\modelinput}{\mathbf{x}}
\newcommand{\modeloutput}{\mathbf{y}}

\newcommand{\modelaction}{\mathbf{a}}
\newcommand{\modelactions}{\mathbf{A}}
\newcommand{\intrain}{\mathbf{\modelinput}}

\newcommand{\intest}{\modelinput^*}

\newcommand{\vaeinput}{\modelinput}  
\newcommand{\vaeencoded}{\mathbf{z}}

\title{\LARGE \bf
Robustness to Out-of-Distribution Inputs\\via Task-Aware Generative Uncertainty
}

\author{Rowan McAllister$^{1}$, Gregory Kahn$^{1}$, Jeff Clune$^{2}$, Sergey Levine$^{1}$%
\thanks{$^{1}$Berkeley AI Research (BAIR), University of California, Berkeley}%
\thanks{${^2}$Uber AI Labs}
}

\begin{document}
\maketitle


\begin{abstract}
Deep learning provides a powerful tool for machine perception when the observations resemble the training data. However, real-world robotic systems, especially mobile robots or autonomous vehicles, must react intelligently to their observations even in unexpected circumstances. This requires a system to reason about its own uncertainty given unfamiliar, out-of-distribution observations. Approximate Bayesian approaches are commonly used to estimate uncertainty for neural network predictions, but can struggle with out-of-distribution observations. Generative models can in principle detect out-of-distribution observations as those with a low estimated density. However, the mere presence of an out-of-distribution input does not by itself indicate an unsafe situation. Intuitively, we would like a perception system that can detect when task-salient parts of the image are unfamiliar or uncertain, while ignoring task-irrelevant features. 
In this paper, we present a method for uncertainty-aware robotic perception that combines generative modeling and model uncertainty to cope with uncertainty stemming from out-of-distribution states, undersampling, and noisy data. Our method estimates an uncertainty measure about the model's prediction, taking into account an explicit (generative) model of the observation distribution to handle out-of-distribution inputs. This is accomplished by probabilistically projecting observations onto the training distribution, such that out-of-distribution inputs map to uncertain in-distribution observations, which in turn produce uncertain task-related predictions, but only if task-relevant parts of the image change. For example, a change of wall color should not confuse a ground robot, while an unfamiliar obstacle should trigger an increase in collision prediction uncertainty.
We evaluate our method on an action-conditioned collision prediction task with both simulated and real data, and demonstrate that our method of projecting out-of-distribution observations improves the performance of four standard Bayesian and non-Bayesian neural network approaches, offering more favorable trade-offs between the proportion of time a robot can remain autonomous and the proportion of impending crashes successfully avoided.
\end{abstract}


\section{Introduction}
\label{sect:introduction}

Deep learning is effective at processing complex sensory input and thus provides a powerful tool for robots, self-driving cars, and other autonomous systems to interpret complex unstructured environments and make intelligent decisions. However, deep neural networks are unable to reliably estimate the confidence (uncertainty) of their own predictions. This poses a major challenge for the deployment of learning-based perception and control systems in safety-critical application areas: when an autonomous system encounters a previously unseen event or obstacle, it should be able to introspectively analyze the confidence of its own predictions and detect situations where it may err, and either take corrective actions or inform a human operator.

A common approach for learning models that produce uncertainty estimates about their predictions is to use approximate Bayesian models, such as Bayesian neural networks~\cite{gal2016uncertainty, mackay1992bayesian, neal2012bayesian}.
A natural question then arises: can Bayesian deep models address the safety challenge in learning-based autonomous systems? While a number of works have proposed that Bayesian neural networks, or similar approaches, can effectively bridge this gap~\cite{gal2017concrete,kahn2017uncertainty}, these approaches do not explicitly account for the input training distribution: powerful discriminative models that rely on approximate Bayesian inference still produce overconfident and poorly calibrated predictions on out-of-distribution data in practice \cite{malinin2018predictive}.

A separate class of methods use generative models to detect out-of-distribution observations~\cite{richter2017safe,schlegl2017unsupervised}. However, out-of-distribution observations alone do not necessarily indicate unsafe situations. For example, a ground robot that uses camera images to avoid collisions using a model trained in buildings with white ceilings might still perform well in buildings with black ceilings.
But if the robot is placed in such a building, generative modeling approaches can immediately detect high uncertainty, even though the color of the ceiling has no effect on model's task. In short, methods based only on detecting out-of-distribution observations are overly pessimistic, being unaware about which parts of the observation are pertinent to the robot's task or safety.

This paper presents a probabilistic framework to handle out-of-distribution states without being overly pessimistic. Our framework uses a generative model to project the input states onto the training distribution, and an action-conditioned Bayesian predictive model that maps the projected input states to task-relevant predictions. Both parts of this framework can be instantiated as deep neural network models that can scale to high-dimensional inputs, such as images. As our case study, we evaluate on action-conditioned collision predictions, where a ground robot must predict, based on an image observation, whether a given sequence of actions will result in a collision. We conduct experiments both in simulation and with real-world data from a remote-control car. Although our method is motivated by the probabilistic structure of the problem, it does not provide concrete guarantees on realistic out-of-distribution inputs when combined with neural network function approximation and approximate inference (and, indeed, we are not aware of any method that does). However, we find that in practice it provides a substantial increase in the ability to avoid catastrophic failures in novel environments with out-of-distribution inputs, when compared to prior methods for uncertainty estimation.


\section{Preliminaries and Motivation}

In this work, we study techniques for uncertainty estimation for deep neural networks in the context of robotic control. The method we propose can be applied to any model that makes predictions about some output variable $\modeloutput$ conditioned on an input $\modelinput$, with weights $\modelparams$. Such a model defines a conditional probability distribution $p(\modeloutput | \modelinput,\modelparams)$.

In a mobile robotics application with vision, $\modelinput$ might correspond to an image observation, and $\modelparams$ would correspond to the weights in the model, which might be, for example, a deep convolutional neural network. The output $\modeloutput$ can be any variable of interest, such as an object detection or future event. Our working example in this paper will be the case where $\modeloutput$ corresponds to a prediction that the robot will collide within a future time interval, although other predictions are compatible with our approach.

Since predictions about future events (including collisions) depend on the course of action that the robot will take, we also condition the model's predictions on the sequence of actions within the time interval between the present and the prediction point, which we denote as $\modelactions$. We can therefore write the action-conditioned prediction model as $p(\modeloutput | \modelinput, \modelactions, \modelparams)$. The particular architecture for this model that we use in our experiments follows prior work~\cite{kahn2017self}, but the derivation is generic to any architecture.

With standard discriminatively trained models, such as deep networks, the most common way to obtain predictions is to learn the maximum a posteriori (MAP) parameter vector $\modelparams_{\text{MAP}}$ by maximizing the log-likelihood: \mbox{$\modelparams_{\text{MAP}} = \argmax_{\modelparams} \sum_i \log p(\modeloutput_i | \modelinput_i, \modelactions_i, \modelparams)$}, summing over all training points $(\modelinput_i, \modelactions_i, \modeloutput_i) \in \datatrain$.
Although the MAP estimate can model uncertainty in the data, it does not capture
model-uncertainty (i.e., uncertainty about the choice of model parameters), which reduces as more data are gathered. This model parameter uncertainty is large when the number of training samples is small relative to the number of parameters, because many different parameter settings can fit the data, and therefore there is uncertainty about which parameter setting is correct. 
To capture model uncertainty, we can adopt a Bayesian approach, where predictions are made by integrating out the parameters according to
\begin{equation}
p(\modeloutput | \modelinput, \modelactions) = \int p(\modeloutput | \modelinput, \modelactions, \modelparams) p(\modelparams | \datatrain) \der\modelparams. \label{eq:bayesian}
\end{equation}
While this integral is typically intractable, it can be approximated in a variety of ways \cite{gal2017concrete,blundell2015weight,efron1994introduction}
to provide an estimate of the model uncertainty.

Although approximate Bayesian methods can address uncertainty stemming from lack of data, they are often less calibrated at data points drawn from outside the training distribution. In this work, we focus on how such models can handle out-of-distribution inputs: unfamiliar situations that deviate systematically from those that the robot saw during training, and which may cause adverse and unpredictable outcomes. This particular type of model-uncertainty is referred to as distributional uncertainty \cite{malinin2018predictive}.
More formally, we will denote the out-of-distribution query as $p(\modeloutput|\intest,\modelactions)$,
where $(\intest,\modelactions,\modeloutput) \sim \ptest(\modelinput,\modelactions,\modeloutput)$, and the distributions $\ptest \neq \ptrain$ have negligible overlap.\footnote{We could instead model the joint $\ptrain(\modelinput,\modelactions)$, but since $\modelactions$ are chosen by the controller or planner, we assume that they are always in-distribution, i.e., $\ptrain(\modelactions) = \ptest(\modelactions)$. However, our method would be straightforward to extend to the case where $\modelactions$ also vary between domains.}
While model-uncertainty can give us a handle on the errors that may be caused by overfitting or small datasets,
it does not necessarily provide us with a reliable mechanism for handling out-of-distribution inputs that are not sampled from $\ptrain$. That is, if the test-time inputs come from $\ptest$, our posterior over the weights $p(\modelparams | \datatrain)$ is often unsuitable. Regardless of how large $\datatrain$ is or how well it represents $\ptrain$, the distributional uncertainty persists.

Intuitively, the problem can be summarized as follows: regardless of how well a Bayesian neural network approach approximates the correct posterior in Equation~\ref{eq:bayesian}, the resulting model only ``knows'' how to accept inputs from $\ptrain$. It can handle inputs sampled from $\ptrain$ that are \emph{not} in $\datatrain$, but inputs from some other distribution $\ptest$ are of the wrong ``type''~\cite{malinin2018predictive}. For example, a vision system trained to drive around in cities may not make sensible predictions on country roads, regardless of how well the Bayesian posterior is approximated; it might therefore make predictions that are overly confident, incorrect, or simply unpredictable. Importantly, the out-of-distribution problem only need be solved insofar as it affects $\modeloutput$, the variable relevant to safety-aware RL. In the next section, we describe our approach for coping with out-of-distribution inputs in the context of safe robot navigation.

\section{Uncertainty with Out-of-Distribution Inputs}
\label{sect:method}

A textbook solution to the problem outlined in the previous section is to simply estimate a new posterior, where the weights are now conditioned on the training set and data from $\ptest$. However, conditioning on $\ptest$ is typically impossible because, by definition, we do not have access to $\ptest$ a priori. An alternative approach would be to adapt to $\ptest$ in an online manner; however, in safety-critical applications such as robot navigation, learning to adapt to $\ptest$ would require actually experiencing potentially catastrophic events, which is most likely unacceptable.

Instead, we adopt an alternative approach, where we aim to explicitly capture and leverage $\ptrain(\modelinput)$. We first summarize our approach intuitively, and then present our model, followed by a practical implementation using deep neural networks.

\subsection{Translating Inputs into the Training Distribution}

As discussed in the previous section, a high-dimensional discriminative predictive model $p(\modeloutput | \modelinput, \modelactions)$ will only make reliable predictions for inputs $\modelinput \sim \ptrain(\modelinput)$. Therefore, when the model is confronted with an out-of-distribution input $\intest \sim \ptest(\modelinput)$, feeding this input to the model may result in an unexpected output for models already trained under $\ptrain$. What if instead we can somehow find an in-distribution image $\modelinput$ that is semantically similar to $\intest$? For example, if the model was trained to drive on city streets and we see an image of a country road, perhaps we can construct an image of a city street with a similar layout. If we can perform this transformation perfectly, the out-of-distribution input problem would be resolved because the model knows what to output for this in-distribution input, and this in-distribution input is semantically similar enough to the out-of-distribution input such that the correct action to take (e.g., where the car should steer) is equivalent. 

Unfortunately, mapping out-of-distribution inputs to become in-distribution is not only difficult from a modeling perspective, but in many cases it might be impossible. Some out-of-distribution inputs may not have any in-distribution analogue that would be meaningful,
and in some cases a single best-guess analogue may result in catastrophe. For example, if an autonomous vehicle believes that a foreign traffic sign indicates go, but in fact the sign indicates stop, the vehicle will likely crash. However, this scenario would happen if we were to only find a single in-distribution analogue. Instead, we can construct a distribution over these analogues, $p(\modelinput | \intest)$, which reflects our uncertainty about which in-distribution inputs best correspond to the out-of-distribution input. If the out-of-distribution input is completely unfamiliar, this distribution should be broad, resulting in highly uncertain predictions that would cause a risk-averse planner or controller to adopt a cautious strategy.

Modeling $p(\modelinput|\intest)$ directly is difficult, especially since $\intest$ is not available during training. Instead, we propose to train a latent variable generative model, which we instantiate in \sect{nn-implementation}, that models the inputs according to
\begin{align}
\vaeencoded \sim p(\vaeencoded), \hspace{0.3in} \modelinput \sim p(\modelinput | \vaeencoded), \hspace{0.3in} \intest \sim p(\intest | \vaeencoded), \label{eqn:distributions}
\end{align}
and models $p(\modelinput | \intest)$ as follows:
\begin{align}
    p(\modelinput | \intest) &= \int p(\modelinput, \vaeencoded | \intest) d\vaeencoded \nonumber \\
    &= \int p(\modelinput | \vaeencoded, \intest) p(\vaeencoded | \intest) d\vaeencoded \nonumber \\
    &= \int p(\modelinput | \vaeencoded) p(\vaeencoded | \intest) d\vaeencoded, \label{eqn:ci} 
\end{align}
where we assume that the latent factor $\vaeencoded$ renders $\modelinput$ and $\intest$ conditionally independent.

Given we do not have access to $\intest\sim\ptest$ at training time, learning $p(\modelinput | \intest)$ from training data is impossible. We therefore assume that $p(\vaeencoded | \intest) = p(\vaeencoded | \modelinput)$, which allows us to train a common encoder model $q(\vaeencoded | \cdot)$ using only $\modelinput \sim \ptrain$. This modeling decision corresponds to the assumption that both domains $\ptrain$ and $\ptest$ have the same underlying factors of variation $\vaeencoded$ (e.g., the position of the road and locations of obstacles). Inferring these factors of variation $\vaeencoded$ from images may thus be done via a common encoder function $q(\vaeencoded | \cdot)$ for both domains, even when the reconstruction probability of the test image $p(\intest | \vaeencoded)$ is unknown and different to $p(\modelinput | \vaeencoded)$.
Note we do not assume the converse, that $p(\intest | \vaeencoded) = p(\modelinput | \vaeencoded)$, since this would imply $p(\intest, \vaeencoded) = p(\modelinput, \vaeencoded)$ which implies $p(\intest) = p(\modelinput)$, violating our original assumption that $\intest$ is out-of-distribution.

\subsection{Uncertainty Estimation}

A natural question to ask at this point is: why don't we simply use the difference between $\modelinput \sim p(\modelinput|\intest)$ and $\intest$ as a measure of uncertainty? This approach resembles methods proposed in prior work~\cite{richter2017safe}, which use reconstruction accuracy (an approximation of observation likelihood) to detect out-of-distribution samples. This metric gives us an approximate notion of how far $\intest$ is from our training images, but it does not actually capture uncertainty about the variable we care about, which is $\modeloutput$. For example, if $\intest$ differs from all in-distribution images in a functionally irrelevant way (e.g., the ceiling is the wrong color), it has high uncertainty, but all reconstructed in-distribution images $\modelinput$ are often functionally the same,
even if they are visually different, resulting in low uncertainty about the variable of interest $\modeloutput$. We observed this phenomenon empirically wherein the reconstructions have different colors, but do not produce different predictions about collisions (seen later in \fig{reconsim}). As we will see in our experiments, simply using reconstruction accuracy tends to produce a poor tradeoff between safety and autonomy.

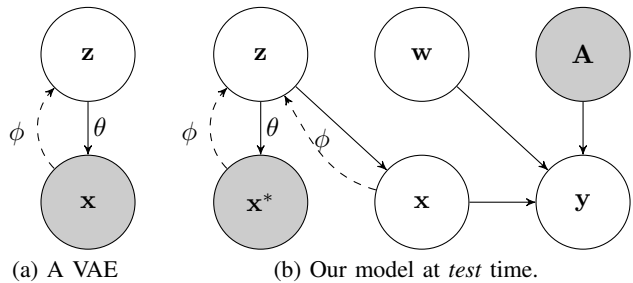
\begin{figure}[b]
    \vspace*{-13pt}
    \centering
    \begin{subfigure}[b]{0.2\columnwidth}
\centering
\begin{tikzpicture}[->,>=stealth',scale=1, transform shape]
\node [matrix,matrix anchor=mid, column sep=25pt, row sep=20pt, ampersand
replacement=\&,nodes={circle,draw,minimum size=1.25cm}] {
\node               (z)   {$\vaeencoded$}; \\
\node[fill=lgrey]   (x)   {$\intrain$}; \\
};
\draw [->] (z) to (x) node [yshift=10mm, xshift=1.6mm]  {$\vaedecoder$};
\draw [dashed,->] (x) to [out=135, in=225] (z) node [yshift=-6mm, xshift=-5mm]  {$\vaeencoder$};
\end{tikzpicture}
\caption{A VAE}
        \label{fig:tikzvae}
    \end{subfigure}
    \hfill
    \begin{subfigure}[b]{0.75\columnwidth}
\centering
\begin{tikzpicture}[->,>=stealth',scale=1, transform shape]
\node [matrix,matrix anchor=mid, column sep=25pt, row sep=20pt, ampersand
replacement=\&,nodes={circle,draw,minimum size=1.25cm}] {
\node[fill=white]           (z)    {$\vaeencoded$}; \&
\node[fill=white]           (w)    {$\modelparams$}; \&
\node[fill=lgrey]           (a)    {$\modelactions$}; \\
\node[fill=lgrey]           (o)    {$\vaeinput^*$}; \&
\node[fill=white]           (i)    {$\vaeinput$}; \&
\node[fill=white]           (c)    {$\modeloutput$}; \\
};
\draw [->] (z) to (o) node [yshift=10mm, xshift=1.6mm]  {$\vaedecoder$};
\draw [->] (z) to (i);
\draw [dashed,->] (i) to [out=165, in=300] (z) node [yshift=-6mm, xshift=5mm]  {$\vaeencoder$};
\draw [dashed,->] (o) to [out=135, in=225] (z) node [yshift=-6mm, xshift=-5mm]  {$\vaeencoder$};
\draw [->] (w) to (c) ;
\draw [->] (a) to (c) ;
\draw [->] (i) to (c) ;
\end{tikzpicture}
\caption{Our model at \textit{test} time.}

        \label{fig:tikzourmodel}
    \end{subfigure}
    \caption{Graphical models. \ref{fig:tikzvae}: A vanilla VAE with encoder $\vaeencoder$ and decoder $\vaedecoder$. \ref{fig:tikzourmodel}: Our probabilistic model where $\modeloutput$ is a existing function of in-distribution images $\modelinput$. We first train our VAE in isolation according to \ref{fig:tikzvae}, however we use it according to \ref{fig:tikzourmodel}. The inference problem we are solving is $p(\modeloutput|\intest)$ which we do by approximately integrating out z, $\intrain$, and $\modelparams$.}
    \label{fig:graphical_model_comaprison}
\end{figure}

Rather than attempting to classify whether an input image is out-of-distribution or not, we instead focus on handling out-of-distribution inputs by only considering the effects relevant to robot safety, namely collision probabilities $\modeloutput$. We
consider this effect by computing the distribution

\vspace{-9pt}
{\footnotesize
\begin{equation}
p(\modeloutput | \intest, \modelactions) = \iiint p(\modeloutput | \modelinput, \modelactions, \modelparams) p(\modelparams | \datatrain) p(\modelinput | \vaeencoded) p(\vaeencoded | \intest) \,\der\modelparams \,\der\modelinput \,\der\vaeencoded.\label{eqn:our_integral}
\end{equation}
}
\vspace{-13pt}

While this integral is generally intractable, we can approximate the inner integral with respect to $\modelparams$ using the standard Bayesian neural network methods in \sect{experimental-results}, and the other two integrals with sampling, where $p(\vaeencoded | \intest)$ is obtained by performing inference in our generative model. Given this approximation for tractability, we cannot guarantee an accurate posterior, and therefore any risk-averse decisions must be tested empirically.
We illustrate our approach in \fig{graphical_model_comaprison}. Here, we assume the same latent factors of variation $\vaeencoded$ give rise to both $\intrain$ and $\intest$, and the prediction model takes in $\modelinput$, $\modelactions$, and $\modelparams$ and predicts $\modeloutput$.

\subsection{Implementation with Deep Neural Networks}\label{sect:nn-implementation}

\begin{figure}[t]
    \centering
    \includegraphics[width=\columnwidth]{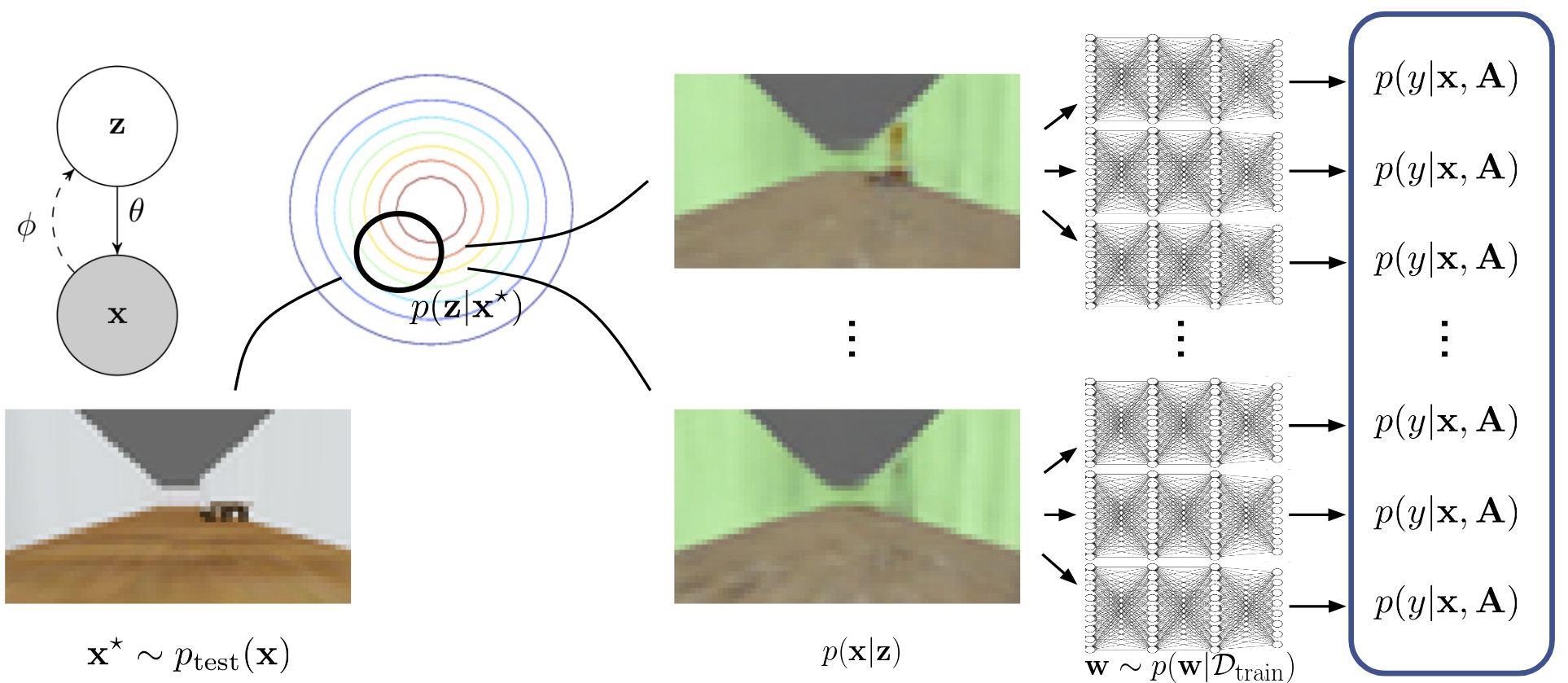}
    \caption{Illustration of the inference procedure for our method at test time. First, a (potentially out-of-distribution) test image $\intest$ is encoded to latent $\vaeencoded$ using the learned VAE model. The latent $\vaeencoded$ is then sampled multiple times and decoded back into images $\modelinput$. The decoded images are now likely to be in-distribution because the VAE was trained to reconstruct in-distribution data. Next, these images are fed through a predictive Bayesian neural network with random weights $\modelparams$ to produce sampled estimates of the predicted output $\modeloutput$. The distribution of sampled $\modeloutput$ can then serve as an estimate of the novelty of (and thus uncertainty about) the input image $\intest$, whose distribution can be used by a risk-averse decision maker.
    }
    \label{fig:ourmodel}
    \vspace*{-5pt}
\end{figure}

We can implement our approach in practice by using a deep neural network generative model. The particular model we use in our implementation is based on the variational autoencoder (VAE)~\cite{kingma2013auto,rezende2014stochastic},
which provides a scalable and efficient way to model high-dimensional observations such as images (see \fig{ourmodel}). A VAE models observations $\modelinput$ as being generated from a low-dimensional latent variable $\vaeencoded$, trained using approximate Bayesian variational inference that maximizes the following lower bound on the marginal log-likelihood:

\vspace{-15pt}
{\small
\begin{align}
\mathcal{L}(\vaedecoder,\vaeencoder;\intrain)\!=\!\Exp{\vaeencoded \sim q_\vaeencoder(\vaeencoded|\intrain)\!}{\log p_\vaedecoder(\intrain|\vaeencoded)}\!-\! \kl{q_\vaeencoder(\vaeencoded|\intrain)}{p(\vaeencoded)}. \label{eqn:vae}
\end{align}
}%
The first term in \eqn{vae} is the model fit, whilst the second is a regularizer that restricts the approximate posterior of $\vaeencoded$ to be close to its prior $p(\vaeencoded)=\mathcal{N}(0,I)$ measured by the Kullback-Leibler divergence. We train a standalone VAE model on in-distribution data during training (\fig{tikzvae}), and our predictive model $p(\modeloutput|\modelinput,\modelactions)$ corresponds to doing inference at test time with the generative model (\fig{tikzourmodel}). Note that approximate inference in the VAE uses the recognition network $q_\vaeencoder(\vaeencoded|\intrain)$, which may be a source of additional error on in-distribution samples, since this network itself might produce unexpected outputs. We found that that this was not a problem in practice, but in principle alternative approximate inference methods, such as sampling, could alleviate problems that arise due to this inference method.

At test time, our VAE also provides us with a convenient way approximately infer $p(\vaeencoded | \modelinput)$ and $p(\vaeencoded | \intest)$ by using the common recognition network $q(\vaeencoded | \cdot)$ trained on $\modelinput \sim \ptrain$. Note using the same function may in practice be problematic, since the network $q$ itself might susceptible to unpredictable behavior for out-of-distribution inputs $\intest$.
Because we do not have formal guarantees, the method must be tested empirically, which we did and we found it to work better than previous methods.
From the information-bottleneck viewpoint of VAEs \cite[appendix B]{alemi2016deep}, we can expect $q_\vaeencoder(\vaeencoded | \intest)$ to `filter out' factors of $\intest$ that have small variation under $\ptrain$, and that $p_\vaedecoder(\modelinput | \vaeencoded)$ will `replace' those missing factors from $\ptrain$.

\subsection{Algorithm Summary}

Our complete method consists of first training a Bayesian neural network model for $p(\modeloutput|\modelinput,\modelactions)$ and a VAE generative model to approximate $p(\modelinput | \vaeencoded)$ and $p(\vaeencoded|\modelinput)$.
Our convolutional VAE architecture uses three layers of 5x5 filters with 32/64/128 filters with strides of 2 with ReLU activations, 32 latents, and a symmetric deconvolutional decoder. Our predictive model receives the reconstructed image and applies four convolutional layers of 8x8/4x4/3x3/3x3 filters of strides 4/2/2/2 before a fully connected layer of 256, which is combined with the action sequence before 16 LSTM layers, then two fully connected layers of size 16, and a scalar softmax output. At training time, the 16 predictions are fit to binary collision events over the following 16 time steps using a softmax cross entropy loss.

At test time, when we observe a new input $(\intest, \modelactions)$, we encode $\intest$ by using $q(\vaeencoded | \intest)$ to obtain a distribution over latent variables, sample from this distribution, and for each latent sample, sample an in-distribution image according to $p(\modelinput|\vaeencoded)$,
the VAE generative model.
We then pass each of these sampled images through the predictive Bayesian neural network to obtain a distribution over predictions for $\modeloutput$, which corresponds to an approximate solution to the triple integral 
in \eqn{our_integral}. From this distribution, we extract a mean and a variance estimate of the time to collision to perform risk-averse decision-making.


\section{Related Work}
\label{sect:related-work}

Prior work has investigated improving the calibration of deep models. Bayesian neural networks based on variational inference have been widely applied to neural network training~\cite{gal2017concrete,blundell2015weight}.
Bootstrapping provides an effective alternative to variational Bayesian methods~\cite{osband2016deep,lakshminarayanan2017simple}, and simple ensembles (without dataset resampling) typically perform just as well as full bootstrapping with deep neural networks~\cite{osband2016deep,lakshminarayanan2017simple}. Indeed, in our experiments, ensembles provide the best uncertainty estimates compared to other Bayesian neural network methods, though our approach improves on all of them. However, while these methods estimate a posterior distribution over the model parameters, they do not explicitly reason about the data distribution itself.

Prior work has also investigated detection of out-of-distribution inputs for learned models~\cite{pimentel2014review}, including through the use of generative models that estimate whether an input can be reconstructed accurately~\cite{richter2017safe,schlegl2017unsupervised,malinin2018predictive}. Such methods typically react simply to an image being out-of-distribution (e.g., by taking emergency action), without regard for whether the discrepancy is salient to the task, while our method uses generative models to estimate predictive uncertainty about a quantity of interest, such as the probability of collision. Prior methods have also explicitly considered uncertainty for action-conditioned models~\cite{kahn2017uncertainty}, but without explicit handling of out-of-distribution inputs.

Our approach aims to map observations into the training distribution, which is similar to the goals of domain adaptation methods that transform target domain images into the source domain, and vice versa, typically for simulation to real world transfer \cite{zhang2018vr,xia2018gibson,hoffman2017cycada}. However, these prior methods do not explicitly reason about uncertainty, and typically employ generative adversarial network models that are known to be poorly calibrated.


\section{Experimental Results}
\label{sect:experimental-results}

Our experiments aim to evaluate how well our approach enables successful risk-averse decision making. We consider a small-scale ground robot attempting to navigate an indoor environment without collisions. Since our method requires a Bayesian neural network component for the predictive model $p(\modeloutput | \modelinput, \modelactions)$, we evaluate it together with several choices of this model. We also compare it to prior Bayesian neural network models that do not use a generative component.

For the predictive model, we evaluate our method with bootstrap ensembles~\cite{efron1994introduction}, concrete dropout~\cite{gal2017concrete} and Bayes-by-backprop~\cite{blundell2015weight}, as well as a standard deterministic neural network model that outputs the probability of collision, which acts as a baseline. Note the Bayes-by-backprop has two variants: Gaussian and Scale Mixture. In our experiments we found Gaussian performed consistently better than Scale Mixture, and we only show the Gaussian variant. For each of these methods, we compare our approach with VAE observation resampling to a baseline that directly uses the predictive model only, which corresponds to the prior Bayesian neural network approaches for uncertainty estimation~\cite{gal2016uncertainty}. We also compare to a purely generative method that only uses the VAE's negative log likelihood (NLL), and triggers interventions whenever it detects out-of-distribution inputs, regardless of their estimated collision probability. Such a method has previously been proposed for safe navigation~\cite{richter2017safe}.

We evaluate each model on its ability to avert collisions on out-of-distribution inputs. If the model predicts that the robot may crash within 2 seconds, the model asks for human intervention. We assume that this will always prevent a collision, but decreases the fraction of time the vehicle is autonomous. An effective method will intervene consistently, but only when there is a non-negligible chance of a collision. We measure the performance by comparing the fraction of crashes avoided versus the fraction of time the vehicle is autonomous, which is very similar to a receiver operating curve (ROC) curve.\footnote{ROC plots show the trade off between the true positive rate (TPR) and false positive rate (FPR) over a set of binary decisions thresholds $\beta$. Our setting is analogous given our binary decision to stop (or not) when predicting that a crash may (or may not) occur: stopping before a crash is a true positive (TP), unnecessary stopping is a false positive (FP), crashing is a false negative (FN), and moving without crashing is a true negative (TN). So our `proportion of crashes avoided' $\equiv$ TPR = TP/(TP+FN), and our `proportion of time not autonomous' $\equiv$ FPR=FP/(FP+TP). Because our stopping decision affects \textit{multiple} time steps, this is not a true ROC curve.} For uncertainty-aware models, including our method, prior Bayesian neural networks, and prior Bayesian models with a VAE, an intervention is requested when $\mu - \beta_\sigma \cdot \sigma < $ 2 seconds, where $\mu$ and $\sigma$ are the mean and standard deviation of the model's predicted time to collision (calculated by performing Monte Carlo sampling 100 times), and $\beta$ is the threshold variable of the ROC-like curve: for large $\beta_\sigma$, the robot will call for help more often and therefore be less autonomous, but safer; for small $\beta_\sigma$, the opposite is true. For the fully deterministic (non-VAE, non-Bayesian) baseline, we request an intervention when $\mu - \beta_\mu < $ 2 seconds. For the NLL baseline, interventions are requested when the input image's VAE NLL surpasses variable threshold $\beta_{\text{NLL}}$.

\subsection{Simulated Robot Car with Unfamiliar Textures}

The simulated robot car and its environment were created using the Bullet physics simulator~\cite{coumans2013bullet} and using Panda3d~\cite{goslin2004panda3d} to render images. The state $\modelinput \in \mathbb{R}^{64 \times 36 \times 3}$ is a color image from an onboard, front facing camera. The action $\modelaction \in \mathbb{R}^{1}$ is the steering angle from $-30^{\circ}$ to $30^{\circ}$, while the speed is fixed to a constant positive value. As the robot drives through the environment, it receives labels $\modeloutput \in \{0, 1\}$ indicating if it has experienced a collision. To ensure a controlled comparison, the evaluation is conducted on a static dataset collected using a random controller with the same random seed for each experiment.

\newcommand{\rulesep}{\unskip\ \vrule\ }
\newcommand{\subfigurewidth}[0]{0.235\columnwidth} 
\begin{figure}[b]
    \centering
    \begin{subfigure}[b]{\subfigurewidth}
        \caption*{Train 1}
        \includegraphics[width=\columnwidth]{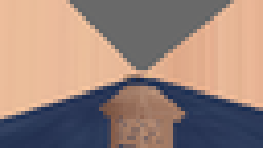}
    \end{subfigure}
    \hfill
    \begin{subfigure}[b]{\subfigurewidth}
        \caption*{Train 2}
        \includegraphics[width=\columnwidth]{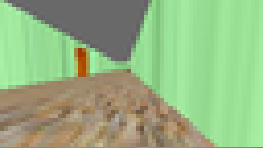}
    \end{subfigure}
    \rulesep
    \begin{subfigure}[b]{\subfigurewidth}
        \caption*{Test 1}
        \includegraphics[width=\columnwidth]{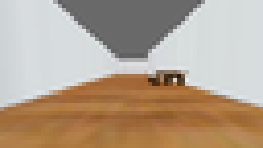}
    \end{subfigure}
    \hfill
    \begin{subfigure}[b]{\subfigurewidth}
        \caption*{Test 2}
        \includegraphics[width=\columnwidth]{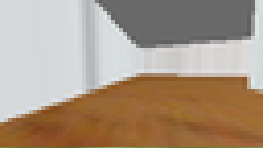}
    \end{subfigure}
    
    \vspace{1mm}
    
    \begin{subfigure}[b]{\subfigurewidth}
        \caption*{reconstructions}
        \includegraphics[width=\columnwidth]{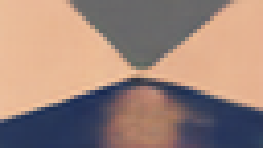}
    \end{subfigure}
    \hfill
    \begin{subfigure}[b]{\subfigurewidth}
        \caption*{reconstructions}
        \includegraphics[width=\columnwidth]{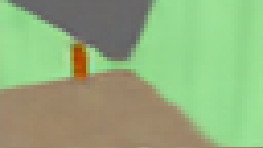}
    \end{subfigure}
    \rulesep
    \begin{subfigure}[b]{\subfigurewidth}
        \caption*{reconstructions}
        \includegraphics[width=\columnwidth]{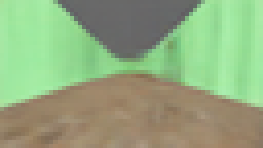}
    \end{subfigure}
    \hfill
    \begin{subfigure}[b]{\subfigurewidth}
        \caption*{reconstructions}
        \includegraphics[width=\columnwidth]{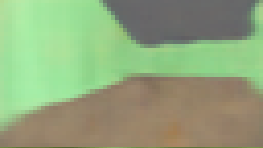}
    \end{subfigure}
    
    \vspace{1mm}
    
    \begin{subfigure}[b]{\subfigurewidth}
        \includegraphics[width=\columnwidth]{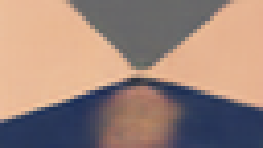}
    \end{subfigure}
    \hfill
    \begin{subfigure}[b]{\subfigurewidth}
        \includegraphics[width=\columnwidth]{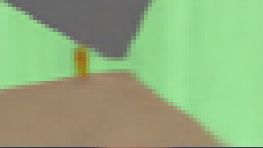}
    \end{subfigure}
    \rulesep
    \begin{subfigure}[b]{\subfigurewidth}
        \includegraphics[width=\columnwidth]{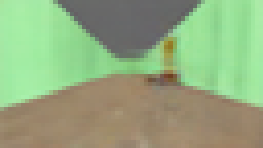}
    \end{subfigure}
    \hfill
    \begin{subfigure}[b]{\subfigurewidth}
        \includegraphics[width=\columnwidth]{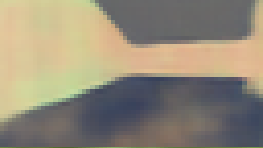}
    \end{subfigure}
    \caption{Simulated robot car observations: the car gathers data in three different training environments and one test environment in which the environment geometry was constant, but the environment textures differed. This figure shows two examples from the training set, and two examples from the test set, each with two sampled VAE reconstructions underneath. Note that multiple VAE reconstructions from a common test image reveal possible uncertainties about object type and color palette. Note that function aspects of the scene can be preserved in at least some of the sampled reconstructions, important for avoiding collisions.}
    \label{fig:reconsim}
\end{figure}

\begin{figure}[t]
    \centering
    \includegraphics[width=\columnwidth]{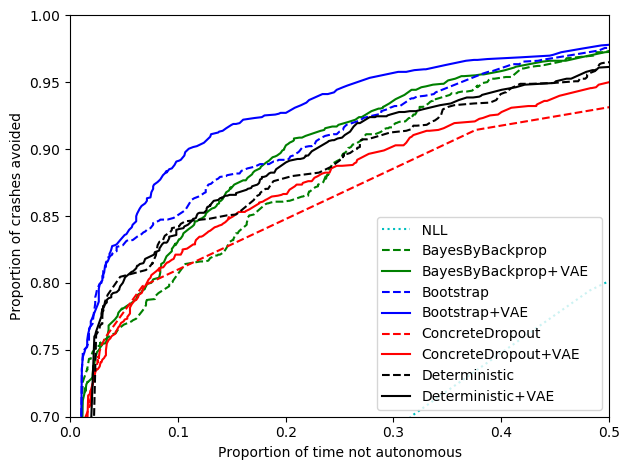}
    \caption{We compare our approach on 30,000 motions from the test environment seen \fig{reconsim} in which the environment textures vary. The robot's goal is to avoid collisions, and can use each method's uncertainty estimation to decide whether to continue driving, or to ask for help. By varying $\beta$, which determines the degree of risk-aversion, each method trades off between safety and autonomy. Prior approaches are shown as dashed lines. Our approach augments each of these prior Bayesian neural network models with the generative component, and each variant of our method is shown with a solid line. In each case, our model improves on the corresponding Bayesian neural network approach.
    The NLL method performs significantly poorer than all other methods, seen in the bottom right.}
    \label{fig:exp-texture-roc}
\end{figure}

We first consider the scenario in \fig{reconsim}. Here, the robot has gathered training data in three environments, and is tested in a new environment with differently textured walls and floors, but with the same environment geometry. Data collected from each environment (including the test environment) comprises 30,000 non-overlapping motions generated from length-16 action sequences $\modelactions$. \fig{exp-texture-roc} shows the ROC-like trade-off for all methods on the test data by varying decision threshold $\beta$. The horizontal axis shows the fraction of time that the robot calls for an intervention, and the vertical axis shows the fraction of crashes that are averted. Methods that are closer to the top left corner of \fig{exp-texture-roc} are preferred because they offer the most favorable trade-off between the proportion of time spent autonomous and the proportion of impending crashes that were successfully avoided. Our approach using VAE tranlations improves the performance of each of the standalone Bayesian neural network approaches---note the gap between the dotted and solid lines of each color---and even improves the performance of the baseline deterministic model. The baseline NLL-only approach generally does not achieve a good tradeoff of autonomy to safety, since the NLL does not contain much task-related information.
This result shows that the generative model component of our approach does capture a notion of uncertainty that is actionable for intelligent decision making for any choice of predictive model after the image transition.

\subsection{Simulated Robot Car with Unfamiliar Obstacles}

\begin{figure}[t]
    \centering
    \begin{subfigure}[b]{0.32\columnwidth}
        \includegraphics[trim=0 110 0 110, clip, width=\columnwidth]{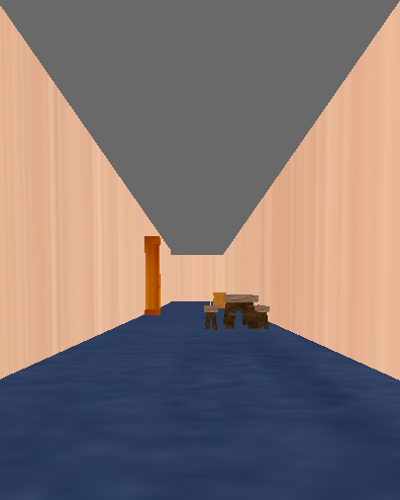}
        \caption{Familiar obstacles}
    \end{subfigure}
    \hfill
    \begin{subfigure}[b]{0.32\columnwidth}
        \includegraphics[trim=0 110 0 110, clip, width=\columnwidth]{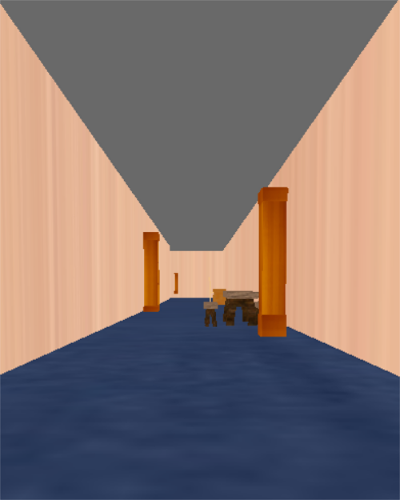}
        \caption{Rearranged}
        \label{fig:exp-cone-familiar}
    \end{subfigure}
    \hfill
    \begin{subfigure}[b]{0.32\columnwidth}
        \includegraphics[trim=0 110 0 110, clip, width=\columnwidth]{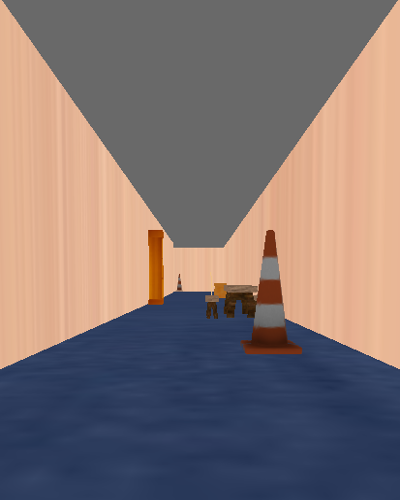}
        \caption{Unfamiliar cones}
        \label{fig:exp-cone-unfamiliar}
    \end{subfigure}
    \caption{The simulated robot car gathered training data in environment (a). The car then gathered data in test environment (b), which consisted of the same set of objects as in (a), but in different locations, and test environment (c), which consisted of some similar objects and novel traffic cone objects.}
    \label{fig:exp-cone}
\vspace{3mm}
    \centering
    \includegraphics[width=\columnwidth]{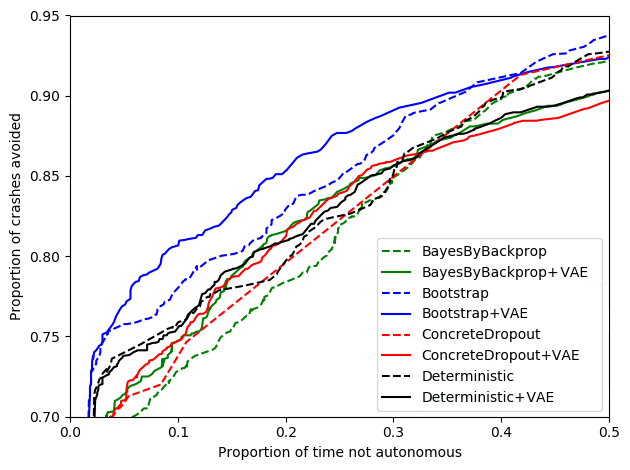}
    \caption{Testing on unfamiliar obstacles (seen \fig{exp-cone-unfamiliar}).
    Our approach again provides better tradeoffs between safety and autonomy compared to each respective Bayesian (see solid line -- ours vs. dashed line -- prior methods).}
    \label{fig:exp-cone-roc}
\vspace{3mm}
    \centering
    \setlength{\unitlength}{1mm}
    \begin{subfigure}[b]{0.025\columnwidth}
        \includegraphics[height=21.5mm,width=\columnwidth]{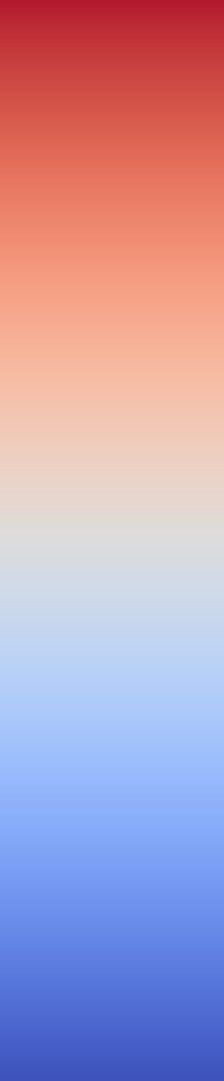}
        \put(0.5,0){\scriptsize{certain}}
        \put(0.5,20){\scriptsize{uncertain}}
        \label{fig:colorscale}
    \end{subfigure}
    \hfill
    \begin{subfigure}[b]{0.47\columnwidth}
        \includegraphics[trim=0 20 0 20, clip,width=\columnwidth]{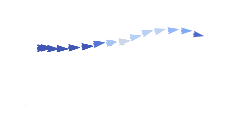}
        \caption{Without cones}
        \label{fig:nocones-bootstrap}
    \end{subfigure}
    \hfill
    \begin{subfigure}[b]{0.47\columnwidth}
        \includegraphics[trim=0 20 0 20, clip,width=\columnwidth]{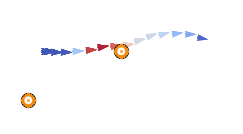}
        \caption{With unfamiliar cones}
        \label{fig:cone-bootstrap}
    \end{subfigure}
    \caption{We qualitatively evaluate the ability of our approach to produce actionable uncertainty estimates in the presence of novel objects not present in the training dataset. (a) and (b) show a top-down view of the car driving through (a) the training environment, which does not contain cones, and (b) the test environment, which does contain novel cones. The uncertainty estimates of our approach in both scenarios are shown by coloring the robot's trajectory, in which blue indicates low uncertainty and red indicates high uncertainty. Our approach produces low uncertainty estimates in the training environment with no cones, but high uncertainty in the test environment when it can see the cone in the camera image.
    }
    \label{fig:cones-bootstrap}
    \vspace{-15pt}
\end{figure}

We next consider the scenario in \fig{exp-cone} in which the robot has gathered training data in an environment containing a specific set of objects, but must then navigate a new environment that also contains an additional previously unseen obstacle -- in this case, the orange and white traffic cones. This scenario differs from the previous one, because now the novel obstacle is not irrelevant: unlike the color of the corridor, the obstacle should \emph{not} be ignored. \fig{exp-cone-roc} shows the ROC curves for all approaches, revealing that our approach improves upon the baseline methods, at least for the regime of interest (when the car is acting more autonomously). \fig{cones-bootstrap} depicts one example the uncertainty estimates produced by our approach, showing that it has low collision uncertainty when in a familiar scene, but high collision uncertainty when novel objects are visible in the camera image.

\subsection{Real Robot Car in an Unfamiliar Building}

\begin{figure}[t!]
    \centering
    \begin{subfigure}[b]{0.24\columnwidth}
        \caption*{Train 1}
        \includegraphics[width=\columnwidth]{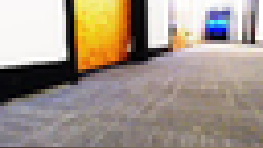}
    \end{subfigure}
    \hfill
    \begin{subfigure}[b]{0.24\columnwidth}
        \caption*{Train 2}
        \includegraphics[width=\columnwidth]{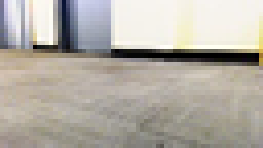}
        
    \end{subfigure}
    \hfill
    \begin{subfigure}[b]{0.24\columnwidth}
        \caption*{Train 3}
        \includegraphics[width=\columnwidth]{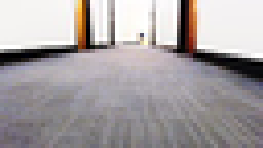}
    \end{subfigure}
    \hfill
    \begin{subfigure}[b]{0.24\columnwidth}
        \caption*{Train 4}
        \includegraphics[width=\columnwidth]{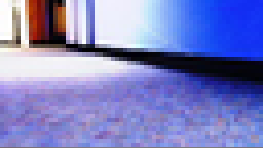}
    \end{subfigure}
    
    \vspace{1mm}
    
    \begin{subfigure}[b]{0.24\columnwidth}
        \caption*{reconstructions}
        \includegraphics[width=\columnwidth]{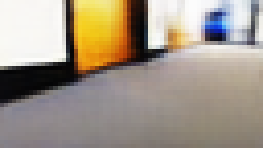}
    \end{subfigure}
    \hfill
    \begin{subfigure}[b]{0.24\columnwidth}
        \caption*{reconstructions}
        \includegraphics[width=\columnwidth]{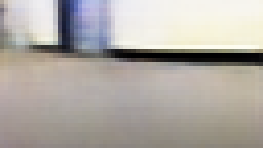}
    \end{subfigure}
    \hfill
    \begin{subfigure}[b]{0.24\columnwidth}
        \caption*{reconstructions}
        \includegraphics[width=\columnwidth]{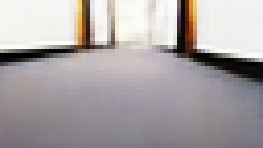}
    \end{subfigure}
    \hfill
    \begin{subfigure}[b]{0.24\columnwidth}
        \caption*{reconstructions}
        \includegraphics[width=\columnwidth]{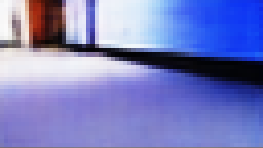}
    \end{subfigure}
    
    \vspace{1mm}
    
    \begin{subfigure}[b]{0.24\columnwidth}
        \includegraphics[width=\columnwidth]{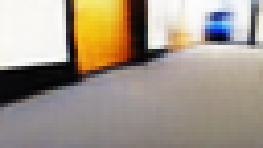}
    \end{subfigure}
    \hfill
    \begin{subfigure}[b]{0.24\columnwidth}
        \includegraphics[width=\columnwidth]{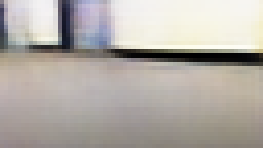}
    \end{subfigure}
    \hfill
    \begin{subfigure}[b]{0.24\columnwidth}
        \includegraphics[width=\columnwidth]{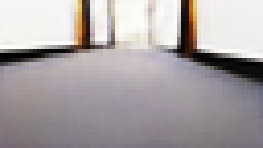}
    \end{subfigure}
    \hfill
    \begin{subfigure}[b]{0.24\columnwidth}
        \includegraphics[width=\columnwidth]{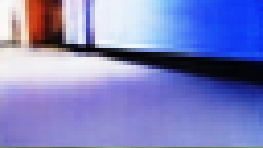}
    \end{subfigure}
    
    \vspace{-2mm}
    \noindent\makebox[\linewidth]{\rule{\columnwidth}{0.4pt}}
    \vspace{-3mm}
    
    \centering
    \begin{subfigure}[b]{0.24\columnwidth}
        \caption*{Test 1}
        \includegraphics[width=\columnwidth]{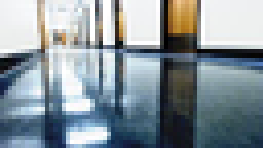}
    \end{subfigure}
    \hfill
    \begin{subfigure}[b]{0.24\columnwidth}
        \caption*{Test 2}
        \includegraphics[width=\columnwidth]{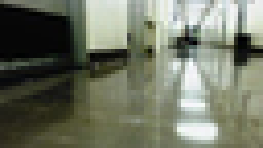}
    \end{subfigure}
    \hfill
    \begin{subfigure}[b]{0.24\columnwidth}
        \caption*{Test 3}
        \includegraphics[width=\columnwidth]{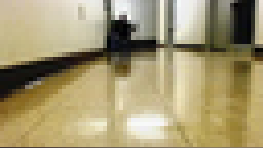}
    \end{subfigure}
    \hfill
    \begin{subfigure}[b]{0.24\columnwidth}
        \caption*{Test 4}
        \includegraphics[width=\columnwidth]{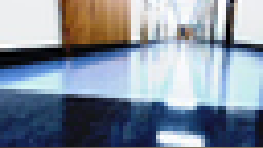}
    \end{subfigure}
    
    \vspace{1mm}
    
    \begin{subfigure}[b]{0.24\columnwidth}
        \caption*{reconstructions}
        \includegraphics[width=\columnwidth]{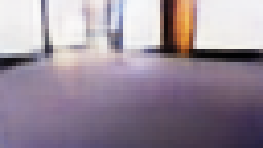}
    \end{subfigure}
    \hfill
    \begin{subfigure}[b]{0.24\columnwidth}
        \caption*{reconstructions}
        \includegraphics[width=\columnwidth]{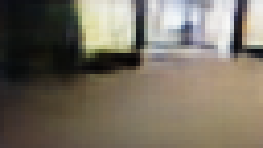}
    \end{subfigure}
    \hfill
    \begin{subfigure}[b]{0.24\columnwidth}
        \caption*{reconstructions}
        \includegraphics[width=\columnwidth]{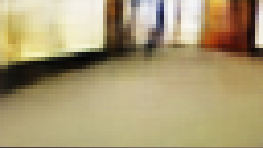}
    \end{subfigure}
    \hfill
    \begin{subfigure}[b]{0.24\columnwidth}
        \caption*{reconstructions}
        \includegraphics[width=\columnwidth]{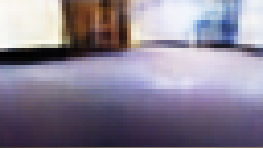}
    \end{subfigure}
    
    \vspace{1mm}
    
    \begin{subfigure}[b]{0.24\columnwidth}
        \includegraphics[width=\columnwidth]{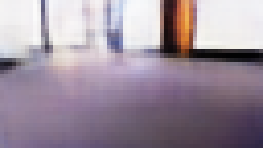}
    \end{subfigure}
    \hfill
    \begin{subfigure}[b]{0.24\columnwidth}
        \includegraphics[width=\columnwidth]{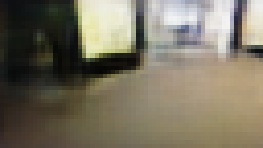}
    \end{subfigure}
    \hfill
    \begin{subfigure}[b]{0.24\columnwidth}
        \includegraphics[width=\columnwidth]{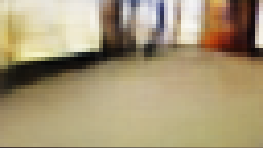}
    \end{subfigure}
    \hfill
    \begin{subfigure}[b]{0.24\columnwidth}
        \includegraphics[width=\columnwidth]{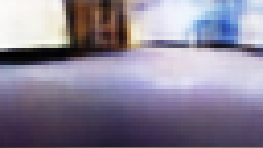}
    \end{subfigure}
    
    \caption{Real robot car observations: the car gathers data in one building to comprise its training set, and a different building for testing. This figure shows four examples from the training set, and four examples from the test set, each with two sampled VAE reconstructions underneath. The reconstruction images for the training set show that the VAE is successfully able to reconstructing training images, while the reconstruction images for the test set show that the VAE is able to successfully transform the novel test input images to look like training images. Note that reconstructions of the test set images often change in ways that do not relate to collision predictions (such as floor color/shininess) but preserve objects, doors, and other things that could cause collisions.}
    \label{fig:reconreal}
    \vspace{-15pt}
\end{figure}

We also evaluated our approach on a real-world dataset gathered from an RC car navigating through multiple buildings. Each building had different hallway widths, and different floor, wall, and ceiling appearances, including both reflective and carpeted floors, as shown in \fig{reconreal}. The training set comprises 53018 non-overlapping motions (one $64 \times 36 \times 3$ image and a length-16 action sequence) in a single building. The test set comprises 29138 motions in a different building with different geometries and textures. The results, shown in \tabl{reconreal}, are qualitatively similar to the simulation results: our method averts more crashes than each of the Bayesian neural network models by themselves for a variety of different autonomy levels. Note there is little difference between methods when assessed on the training data (see appendix), since they use the same network architecture and are equally flexible.

\begin{figure}[t]
    \centering
    \includegraphics[width=\columnwidth]{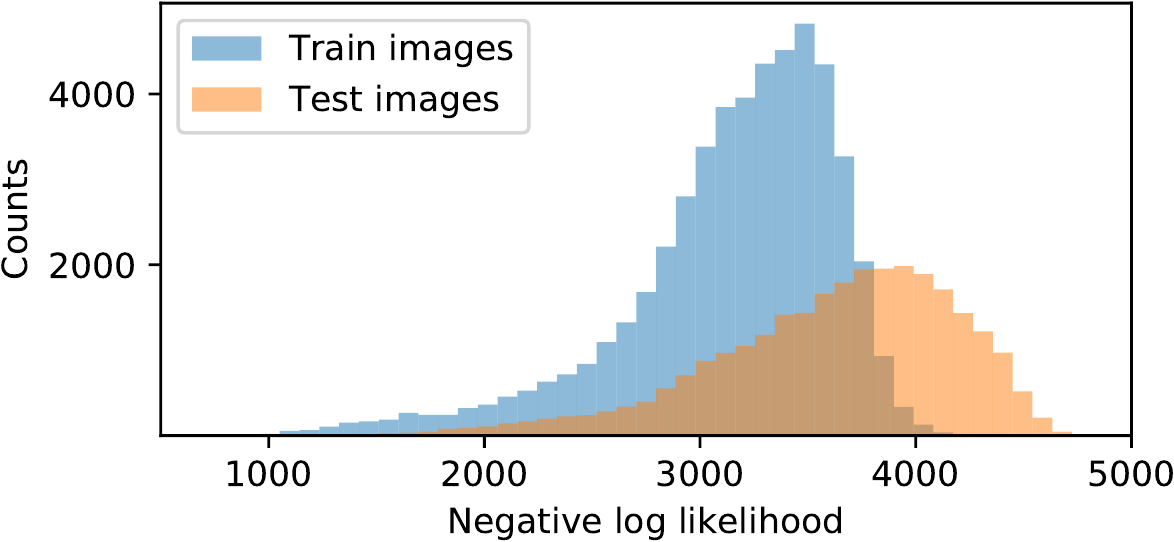}
    \caption{Histogram comparison showing how real training images and real test images compare w.r.t.\ the VAE negative log likelihood (NLL). Whilst in-distribution training data is generally expected to have lower NLL than out--of-distribution test data, their distributions have considerable overlap. Thus, we cannot expect to use NLL information alone to detect out-of-distribution images \cite{nalisnick2018do}, let alone decide an appropriate action given them.}
    \label{fig:nll}
    \vspace{-15pt}
\end{figure}

As an additional baseline, we also combined the deterministic model's stopping rule $\beta_\mu$ with the NLL baseline stopping rule $\beta_{\text{NLL}}$ to create more flexible stopping rules: $\mu - \beta_\mu - \beta_{\text{NLL}} < 2$ seconds, testing all combinations of $\beta_\mu$ and $\beta_{\text{NLL}}$.
We found an image's NLL carries little additional information relevant to the task of avoiding collisions given unusual images at test time. 
Indeed \fig{nll} shows there is considerable overlap between both in-distribution and out-of-distribution image NLLs, meaning it is difficult to even detect (let alone handle) out-of-distribution inputs using NLL information alone. Indeed we found this information is often misleading.
For instance, we recorded all threshold combinations $\beta_\mu$ and $\beta_{\text{NLL}}$ that maximized the ROC curve at each point along the x-axis, and ran the same set of intervention rules to generate the \textit{Deterministic+NLL} data in \tabl{reconreal}. The richer set of combined decision rules perform poorer on out-of-distribution test images than the \textit{Deterministic} baseline alone. Instead of using the VAE for NLL decision rules, using the VAE to reconstruct images is more robust to out-of-distribution images. For instance \textit{Deterministic+VAE} outperforms both other deterministic baselines in \tabl{reconreal}. Our proposed method, \textit{Bootstrap+VAE} generally achieves the best trade-off compared to all other methods.

\begin{table}[t]
\centering
\caption{The percentage of crashes averted per method (rows) for different percentages of time spent autonomous (columns) on the real car in the test environment with 29138 out-of-distribution images. Note the VAE reconstruction helps in almost all cases compared to their non-reconstruction counterparts.
}
\label{tabl:reconreal}
\begin{tabular}{l c c c c c} 
Method & 50\% & 60\% & 70\% & 80\% & 90\% \\
\hline
NLL & 23.6 & 18.2 & 13.3 & 8.8 & 7.6 \\
\rowcolor{Gray}
BayesByBackprop & 83.2 & 79.9 & 76.5 & 69.1 & 56.3 \\
\rowcolor{Gray}
BayesByBackprop + VAE & 89.1 & 85.3 & 79.5 & 70.9 & 56.1 \\
Bootstrap & 89.6 & 86.2 & 81.7 & 75.3 & \textbf{65.2} \\
Bootstrap + VAE & \textbf{91.4} & \textbf{87.8} & \textbf{83.0} & \textbf{76.6} & 64.0 \\
\rowcolor{Gray}
ConcreteDropout & 82.0 & 78.5 & 74.9 & 69.4 & 56.8 \\
\rowcolor{Gray}
ConcreteDropout + VAE & 88.9 & 84.5 & 79.1 & 70.9 & 56.6 \\
Deterministic & 85.3 & 80.0 & 73.9 & 64.9 & 50.7 \\
Deterministic + NLL & 81.3 & 73.3 & 62.0 & 35.5 & 7.6 \\
Deterministic + VAE & 87.5 & 81.8 & 76.1 & 66.8 & 50.8 \\
\hline
\end{tabular}
\vspace{-2pt}
\end{table}

\begin{figure}[t]
    \centering
    \begin{subfigure}[b]{\columnwidth}
        \caption{Mean latent values in tSNE}
        \label{fig:tsne1}
        \includegraphics[width=\columnwidth]{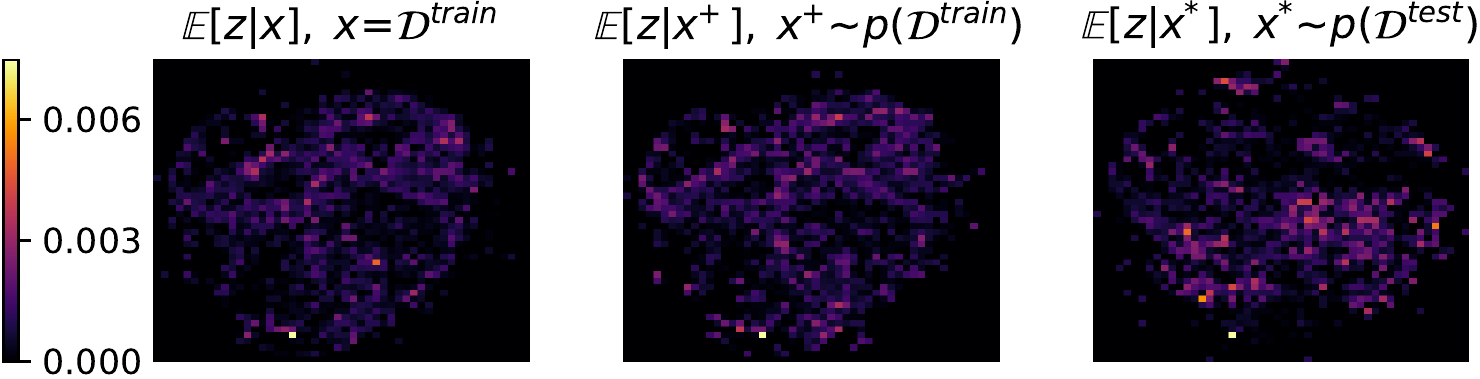}
    \end{subfigure}
    
    \begin{subfigure}[b]{0.1\columnwidth}
        \includegraphics[width=\columnwidth]{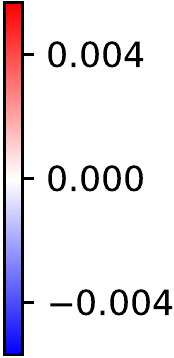}
    \end{subfigure}
    \begin{subfigure}[b]{0.405\columnwidth}
        \caption{Probability difference}
        \label{fig:tsne2}
        \includegraphics[width=\columnwidth]{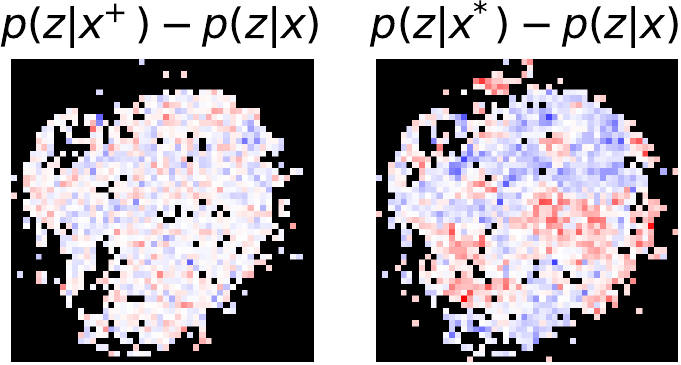}
    \end{subfigure}
    \begin{subfigure}[b]{0.465\columnwidth}
        \caption{Support difference}
        \label{fig:tsne3}
        \includegraphics[width=\columnwidth]{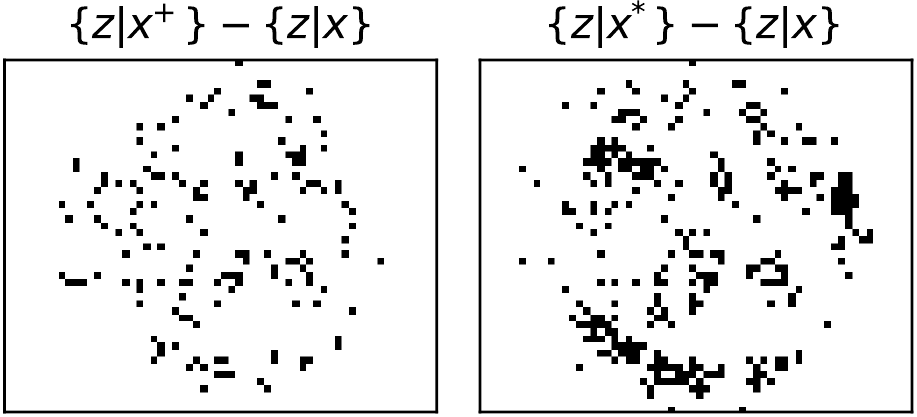}
    \end{subfigure}
    \caption{
    \fig{tsne1} shows the tSNE of the sampled latents for the training data used to train the VAE (left), additional in-distribution data (middle), and out-of-distribution data (right). Note the left and middle densities appear similar, whilst the out-of-distribution data (right) looks different.
    \fig{tsne2} show the differences between the sampled distributions (black means no samples). White regions depict similar densities, while red/blue are show degrees of difference. We again see that out-of-distribution data is more dissimilar than in-distribution data compared to the original training data, showing out-of-distribution images often project to regions of latent space with low density w.r.t.\ training images.
    Finally, \fig{tsne3}, shows the difference in sampled distributions support. This illustrates that even though out-of-distribution images generally have latent values that are unlikely under the latent training distribution, they are still likely to have latent values that also exist in the training set.
    }
    \label{fig:tsne}
    \vspace{-15pt}
\end{figure}

To gain some insight into how out-of-distribution images are encoded, we visualize the latent space $\vaeencoded$ in two dimensions (\fig{tsne}) using t-SNE~\cite{maaten2008visualizing}. To do so, we first computed the latent $\vaeencoded$ Gaussian distributions for both train and test images. Second, for each latent distribution, we drew 10 samples. Third, we inputted all samples into t-SNE, such that t-SNE does not know which samples are train and which are test. Finally, we show the differences of 1) the training data used to train the VAE, 2) additional in-distribution data, and 3) out-of-distribution data. \fig{tsne} shows that even though most out-of-distribution images have latent values that are unlikely under the latent encoding of training images, they rarely have values that were never used by the training distribution. Thus the VAE loss function \eqref{eqn:vae} encountered most out-of-distribution latent values during training, and fit them to decode to a reasonable in-distribution image. 

\section{Conclusion}
\label{sect:conclusion}

We introduced a probabilistic framework that aims to cope with uncertainty stemming from out-of-distribution states, undersampling, and noisy data.
We combined recent advances in generative models 
with model-uncertainty estimation methods to improve the tradeoff between avoiding catastrophic collisions and maintaining a high degree of autonomy
on simulated and real-world robot car navigation datasets.
Our approach to uncertainty estimation via generative modeling can be combined with any existing Bayesian neural network approach, and we found experimentally that it provides an improvement with respect to this tradeoff metrics for all Beysian neural network methods that we evaluated.
While our method empirically outperforms prior techniques on a range of comparisons, our approach also has a number of limitations. The inference procedure at test-time still uses the VAE encoder network, which itself may not be robust to out-of-distribution inputs. This issue could in principle be mitigated by using other approximate inference methods, such as Markov chain Monte Carlo, and practically we observed that our method still produces actionable uncertainty estimates in spite of this limitation. However, developing a more resilient inference methodology would be an exciting direction for future work. Our method also does not provide any \emph{theoretical} guarantee that the model will respond correctly in practice. Indeed, to our knowledge, neither does any other prior method, in the case of neural network models and image observations. Theoretical analysis of uncertainty estimation for such settings is an important direction for future work.



\FloatBarrier
\appendix
\section{Appendix}

We expect all methods should perform approximately equally on the training data, given they are all equally and sufficiently flexible (except the NLL-only method), which we see in \tabl{reconrealtrain}. Indeed we used the same network architecture for each predictive model. By contrast, performance on the \textit{test} set differs (\tabl{reconreal}), not because architecture or flexibility differences, but because of the complimentary benefits of translating out-of-distribution inputs and reasoning about uncertainty in the predictive model.

\begin{table}[h]
\centering
\caption{The percentage of crashes averted per method (rows) for different percentages of time spent autonomous (columns) on the real car in the in-distribution \textbf{training} environment only. This table is not intended to compare methods on what we ultimately care about (generalizing to a test set), merely to show that all methods perform similarly on the training environment.}
\label{tabl:reconrealtrain}
\begin{tabular}{l c c c c c} 
Method & 50\% & 60\% & 70\% & 80\% & 90\% \\
\hline
NLL & 18.3 & 13.7 & 9.7 & 6.1 & 2.9 \\
\rowcolor{Gray}
BayesByBackprop & 95.4 & 94.4 & 93.5 & 92.6 & 90.4 \\ 
\rowcolor{Gray}
BayesByBackprop + VAE & 99.9 & 99.4 & 98.5 & 97.0 & 91.4 \\
Bootstrap & 100.0 & 99.9 & 99.5 & 97.8 & 92.5 \\
Bootstrap + VAE & 99.8 & 99.4 & 98.5 & 96.6 & 93.0 \\
\rowcolor{Gray}
ConcreteDropout & 95.3 & 94.3 & 93.4 & 92.5 & 91.1 \\
\rowcolor{Gray}
ConcreteDropout + VAE & 100.0 & 99.8 & 98.6 & 97.1 & 90.9 \\
Deterministic & 100.0 & 100.0 & 99.6 & 96.0 & 88.7 \\
Deterministic + NLL & 100.0 & 99.9 & 99.5 & 96.0 & 88.7 \\
Deterministic + VAE & 99.9 & 99.5 & 97.5 & 91.9 & 86.0 \\
\hline
\end{tabular}
\end{table}

\end{document}